\documentclass[10pt,twocolumn,letterpaper]{article}
\usepackage{caption}
\pdfoutput=1

\usepackage{3dv}
\usepackage{times}
\usepackage{epsfig}
\usepackage{graphicx}
\usepackage{amsmath}
\usepackage{amssymb}
\usepackage{mathtools}
\usepackage{graphicx}
\usepackage[table,xcdraw]{xcolor}
\usepackage{float}
\DeclarePairedDelimiterX{\infdivx}[2]{(}{)}{%
  #1\;\delimsize\|\;#2%
}
\usepackage{enumitem}
\usepackage{algorithmic,algorithm}
\usepackage{subfig}
\usepackage{appendix}

\newcommand{\infdiv}{\text{KL}\infdivx}

\def\mypar#1{\vspace{0mm}{\noindent\bf #1.}\hspace{1mm}}
\def\mypars#1{\vspace{0mm}{\noindent\bf #1}}

\usepackage[pagebackref=true,breaklinks=true,colorlinks,bookmarks=false]{hyperref}
\def\mb#1{\mathbf{#1}}
\usepackage{multirow}
\def\fig#1{Figure~\ref{fig:#1}}
\def\tab#1{Table~\ref{tab:#1}}
\def\sect#1{Sec.~\ref{sec:#1}}
\def\Eq#1{Eq.~(\ref{eq:#1})}
\def\app#1{Appendix~\ref{sec:#1}}

\threedvfinalcopy
\begin{document}

\title{Stochastic Neural Radiance Fields: \\
Quantifying Uncertainty in Implicit 3D Representations}

\author{Jianxiong Shen, Adria Ruiz, Antonio Agudo, Francesc Moreno-Noguer\\
{Institut de Robòtica i Informàtica Industrial, CSIC-UPC,
Barcelona, Spain}\\
{\tt jianxiong.shen@upc.edu}}

\maketitle

\begin{abstract}
Neural Radiance Fields (NeRF) has become a popular framework for learning implicit 3D representations and addressing different tasks such as novel-view synthesis or depth-map estimation. However, in downstream applications where decisions need to be made based on automatic predictions, it is critical to leverage the confidence associated with the model estimations. Whereas uncertainty quantification is a long-standing problem in Machine Learning, it has been largely overlooked in the recent NeRF literature. In this context, we propose Stochastic Neural Radiance Fields (S-NeRF), a generalization of standard NeRF that learns a probability distribution over all the possible radiance fields modeling the scene. This distribution allows to quantify the uncertainty associated with the scene information provided by the model. S-NeRF optimization is posed as a Bayesian learning problem which is efficiently addressed using the Variational Inference framework. Exhaustive experiments over benchmark datasets demonstrate that S-NeRF is able to provide more reliable predictions and confidence values than generic approaches previously proposed for uncertainty estimation in other domains.
\end{abstract}

\section{Introduction}

Recent learning based methods have shown impressive  results in 3D modeling. In particular, implicit neural volume rendering~\cite{OccupancyNetworks,deepsdf,neuralvolume,nerf} has become a popular framework to learn  compact 3D scene representations from a sparse set of images. 
Among these methods, Neural Radiance Fields (NeRF)~\cite{nerf} has received a lot of attention given its ability to render photo-realistic novel views of the scene. Additionally, several works have shown that the 3D representations learned by NeRF can be used for different downstream tasks, such as camera-pose recovery~\cite{iNeRF}, 3D semantic segmentation~\cite{semantic_srn} or depth estimation~\cite{nerf}. 
Even though all these tasks have relevant applications in fields such as robotics or augmented reality, existing NeRF-based approaches  are limited in these scenarios, for being unable to provide information about the confidence associated with the model outputs. 
For instance, consider a robot using Neural Radiance Fields to reason about its environment. In order to plan the optimal actions and reduce potential risks, the robot must take into account not only the outputs produced by NeRF, but also their associated uncertainty.

In this context, we propose Stochastic-NeRF, a generalization of the original NeRF framework able to quantify the uncertainty associated with the implicit 3D representation. Unlike standard NeRF which only estimates deterministic radiance-density values for all the spatial-locations in the scene, S-NeRF models these pairs as stochastic variables following a distribution whose parameters are optimized during learning. In this manner, our method implicitly encodes a distribution over all the possible radiance fields modeling the scene. The introduction of this stochasticity enables S-NeRF to quantify the uncertainty associated with the the resulting outputs in different tasks such as novel-view rendering or depth-map estimation (see \fig{intro}).
During learning, we follow a Bayesian approach to estimate the posterior distribution of all the possible radiance fields given training data. To make this optimization problem tractable, we devise a learning procedure for S-NeRF based on Variational Inference \cite{blei2017variational}.
Conducting exhaustive experiments over benchmark datasets, we show that S-NeRF is able to provide more reliable uncertainty estimates than generic approaches previously proposed for uncertainty estimation in other domains. In particular, we evaluate the ability of S-NERF to quantify the uncertainty in novel-view synthesis and  depth-map estimation.

\section{Related Work}
\label{sec:related}
\mypar{Neural Radiance Fields} Similar to other neural volumetric approaches such as Scene Representation Networks~\cite{srn} or Neural Volumes~\cite{neuralvolume}, NeRF uses a collection of sparse 2D views to learn a neural network encoding an implicit 3D representation of the scene. NeRF employs a simple yet effective approach where the network predicts the volume density and emitted radiance for any given view-direction and spatial coordinate. These outputs are then combined with volume rendering techniques~\cite{max1995optical}  to synthesize novel views or estimate the implicit 3D geometry of the scene.

Since it was firstly introduced, many works have extended the original NeRF framework  to address some of its  limitations. For instance, ~\cite{Liu20neurips_sparse_nerf,Lindell_AutoInt,neff2021donerf,Lombardi2021MixtureOV} explored several techniques to accelerate the time-consuming training and rendering process. Other works~\cite{nerfvae, pixelnerf} introduced the notion of ``scene priors", allowing a single NeRF model to encode the information of different scenes and generalize to novel ones. Similarly, ~\cite{nerfwild} proposed to account for illumination changes and transient occluders in order to leverage in-the-wild training views. Other recent works~\cite{Pumarola_D_NeRF,Gafni_DNRF,raj2021pva,Li2021_nsff,Xian2021_stnif,peng2021neural} have extended NeRF to cases with dynamic objects in the scene.

Different from the aforementioned methods, the proposed S-NeRF explicitly addresses the problem of estimating the uncertainty associated with the learned implicit 3D representation. Our framework is a probabilistic generalization of original NeRF and thus, our formulation can be easily combined with most of previous works in order to improve different aspects of the model. 


\vspace{1mm}
\mypar{NeRF Applications} Implicit representations learned by NeRF can be used to infer useful scene information for domains such as Robotics or Augmented Reality (AR). For instance, the ability to render novel views, estimate 3D meshes~\cite{nerf,Pumarola_D_NeRF} or recover camera-poses\cite{iNeRF} can be used to allow robots to reason about the environment and plan navigation paths or object manipulations. Additionally, ~\cite{GIRAFFE} proposed to incorporate a compositional 3D scene representation into a generative model to achieve controllable novel view synthesis. This capability is specially interesting in AR scenarios. Despite these potential applications, previous NeRF approaches are still limited in the aforementioned domains. The reason is that they are not able to quantify the underlying uncertainty of the model and thus, it is not possible to evaluate the risk associated with downstream decisions based on the output estimations. 

Recently, NeRF-in-the-Wild (NeRF-W)~\cite{nerfwild} considered to identify the uncertainty produced by transient objects in the scene such as pedestrians or vehicles. In particular, the authors proposed to estimate a value indicating the variance for each rendered pixel in a novel synthetic view. This variance is computed by treating it as an additional value to be rendered analogously to pixel RGB intensities. However, this approach has two critical limitations. Firstly, pixel-colors are produced by a specific physical process that is not related to the model uncertainty. As a consequence, estimating the latter with volume rendering techniques is not theoretically-founded and can lead to sub-optimal results. On the other hand, while NeRF-W is able to predict the variance associated with the rendered pixels, it does not explicitly model the uncertainty of the radiance field representing the scene. Hence, it can not quantify confidence estimates about the underlying 3D geometry.

To the best of our knowledge, S-NeRF is the first approach to explicitly model the uncertainty of the implicit representation learned by NeRF. In contrast to NeRF-W, our method allows us to quantify the uncertainty associated not only with rendered views, but also with estimations related with the 3D geometry (see Fig.~\ref{fig:intro} again).  

\vspace{1mm}
\mypars{Uncertainty Estimation} is a long-standing problem in Deep Learning \cite{reviewuncertainty,guo2017calibration,ProbabilisticBF,mc-dropout,deepensemble}. To address it, a popular approach imposes the Bayesian Learning framework \cite{bayesianTheory} to estimate the posterior distribution over the model given observed data. This posterior distribution can be used during inference to quantify the uncertainty of the model outputs. In the context of deep learning, Bayesian Neural Networks ~\cite{Tran2019BayesianLA,PosteriorNetwork,ProbabilisticBF,Maddox2019ASB} use different strategies to learn the posterior distribution of the network parameters given the training set. However, these approaches are typically computationally expensive and require significant modifications over network architectures and training procedures.

To address this limitation, other approaches have explored other strategies to implicitly learn the parameter distribution. For instance, dropout-based methods  ~\cite{mc-dropout,Aralikatti2018GlobalSE,Hernndez2020ImprovingPU,variational_dropout} introduce stochasticity over the intermediate neurons of the network in order to efficiently encode different possible solutions in the parameter space. By evaluating the model with different dropout configurations over the same input, the uncertainty can be quantified by computing the variance over the set of obtained outputs. A similar strategy consists on using deep ensembles ~\cite{deepensemble,jain2020maximizing}, where a finite set of independent networks are trained and evaluated in order to measure the output variance. Whereas these solutions are more simple and efficient than Bayesian Neural Networks, they still require multiple model evaluations. This limits their application in NeRF, where the rendering process is already computationally expensive for a single model. 

Different from the previous approaches learning a posterior distribution over the model parameters, the proposed S-NeRF learns a single network encoding the distribution over all the possible radiance fields modelling the scene. As we will discuss in the following sections, this allows to efficiently obtain uncertainty estimates without the need of evaluating multiple model instances.

\section{Stochastic Neural Radiance Fields}
\label{sec:s_nerf}

\subsection{From Standard to Stochastic-NeRF}
\label{sec:svsd_nerf}

\vspace{1mm}
\noindent\textbf{Standard NeRF} \cite{nerf} models a 3D volumetric scene as a radiance field $\mathcal{F}$ defining a set:
\begin{equation}
    \label{eq:d_nerf_set}
    \mathcal{F} = \{ (\mathbf{r}(\mathbf{x},\mathbf{d}), \alpha(\mathbf{x})): \mathbf{x} \in \mathbb{R}^3, \mathbf{d} \in \mathbb{R}^2
\}
\end{equation}
where $\alpha(\mathbf{x}) \in \mathbb{R}^+$ is the volume density in a specific 3D spatial-location $\mathbf{x}$  and  $\mathbf{r}(\mathbf{x},\mathbf{d}) \in \mathbb{R}^3$ is the emitted RGB radiance which is also dependent on the view direction $\mathbf{d}$. 

To model the radiance field $\mathcal{F}$, NeRF uses a parametric function $f_{\boldsymbol{\theta}}(\mathbf{x},\mathbf{d}): \mathbb{R}^3 \times \mathbb{R}^2 \rightarrow \mathbb{R}^4$ which encodes the radiance $\mathbf{r}$ and density $\alpha$ for every possible location-view pair $(\mathbf{x},\mathbf{d})$ in the scene . Concretely, this function is implemented by a deep neural network with parameters $\boldsymbol{\theta}$.

\vspace{1mm}
\noindent\textbf{NeRF Optimization}: For a given scene, NeRF optimizes the network $f_{\boldsymbol{\theta}}$ by leveraging a training set $\mathcal{T}=\{(\mathbf{c}^1,\mathbf{x}^1_o,\mathbf{d}^1),\dots,(\mathbf{c}^N,\mathbf{x}^N_o,\mathbf{d}^N) \}$ formed by $N$ triplets, where $\mathbf{c}^n$ is a RGB pixel-color captured by a camera located in a 3D position $\mathbf{x}_o^n$ in the scene. Additionally, $\mathbf{d}^n$ is the normalized direction from the camera origin to the pixel in world-coordinates. This training set can be obtained by capturing a collection of views from the scene using different cameras with known poses. 

By assuming that training samples $(\mathbf{c}^n,\mathbf{x}^n_o,\mathbf{d}^n) \in \mathcal{T}$ are independent observations, the network parameters $\boldsymbol{\theta}$ are optimized by minimizing the negative log-likelihood as:
\begin{align}
    \min_{\boldsymbol{\theta}} \hspace{1mm} -\log p_{\boldsymbol{\theta}}(\mathcal{T}) = - & \frac{1}{N} \sum_{n=1}^N \log p_{\boldsymbol{\theta}}(\mathbf{c}^n | \mathbf{x}^n_o ,\mathbf{d}^n )  \nonumber \\
    \propto& \frac{1}{N} \sum_{n=1}^N ||\mathbf{c}^n-C(\mathbf{x}^n_o,\mathbf{d}^n)||_2^2
\end{align}
where the quadratic error follows from defining $p_{\boldsymbol{\theta}}(\mathbf{c}_n^p | \mathbf{x}^o_n ,\mathbf{d}_n)) = \mathcal{N}(\mathbf{c}_n^p | \mu, \sigma)$ as a Gaussian distribution with unit variance and a mean defined by the volumetric rendering function:

\begin{equation}
\label{eq:vol_rendering}
C(\mathbf{x}_o,\mathbf{d}) = \int_{t_s}^{t_f} \frac{ \mb r_t \alpha_t}{\exp( \int_{t_s}^{t}  {\alpha}_s ds)}  dt
\end{equation}
where $\mathbf{x}_t = \mathbf{x}^o + t \mb d$ is a specific spatial-location along a ray with direction $\mathbf{d}$ which crosses the scene from the pixel position in world-coordinates $\mathbf{x}_{t_s}$ to the point $\mathbf{x}_{t_f}$. Additionally, we express  $\mathbf{r}(\mathbf{x}_t,\mathbf{d})$ and $\alpha(\mathbf{x}_t)$ as $\mb r_t$ and $\alpha_t$, respectively. More details about the volumetric rendering function in \Eq{vol_rendering} and how it is approximated can be found in~\cite{nerf}. 

\vspace{1mm}
\noindent\textbf{Stochastic NeRF} is a generalization of the previously described framework. Specifically, instead of learning a single radiance field $\mathcal{F}$, S-NeRF models a distribution $p(\mathcal{F})$ over all the possible fields modelling the scene. For that purpose, we consider that for each location-view pair $(\mathbf{x},\mathbf{d})$, the volume density and emitted radiance are random variables following an unknown joint distribution. In this manner, any radiance field $\mathcal{F}$ defined in \Eq{d_nerf_set} can be considered a realization over the distribution $p(\mathcal{F})$. As we will discuss in \sect{synth_unc}, treating the radiance field as a set of stochastic variables allows to reason about the underlying uncertainty in the implicit 3D representation. 

\vspace{1mm}
\noindent\textbf{S-NeRF Optimization}: Different from the optimization strategy used in original NeRF, S-NeRF adopts a Bayesian approach where the goal is to estimate the posterior distribution of the possible radiance fields $\mathcal{F}$ given the observed training set $\mathcal{T}$:
\begin{equation}
    \label{eq:snerf_posterior}
    p(\mathcal{F} | \mathcal{T}) = \frac{p( \mathcal{T} | \mathcal{F}) p(\mathcal{F})}{p(\mathcal{T})} 
\end{equation}
where $p( \mathcal{T} | \mathcal{F})$ is the likelihood of $\mathcal{T}$ given a radiance field and $p(\mathcal{F})$ is a  distribution modelling our prior knowledge about the radiance and density pairs over the different spatial-locations in the scene.

\begin{figure*}
\centering
  \includegraphics[width=\textwidth]{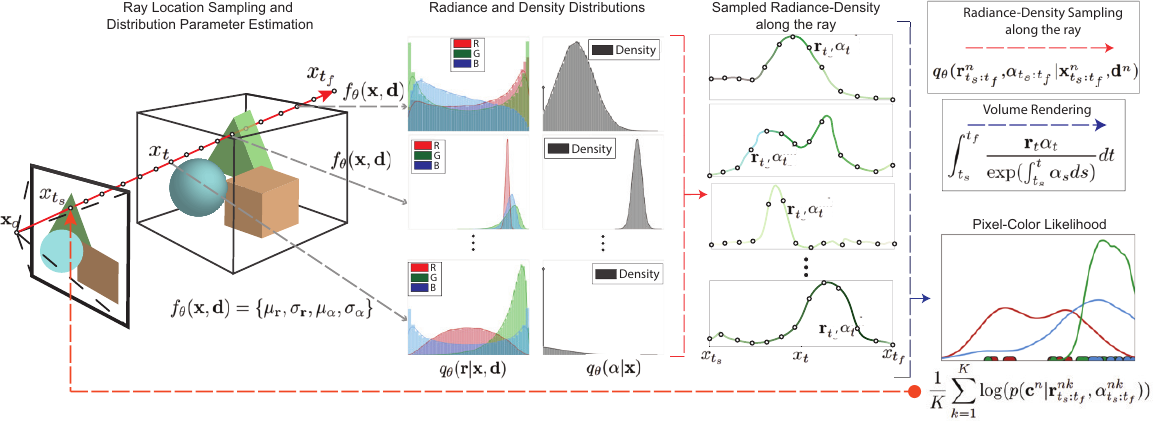}
  \caption{\textbf{Illustration of the pipeline used by S-NeRF to compute the log-likelihood of the pixel color in a given view.} From left to right: (i) For each coordinate $\mathbf{x}$ along a camera ray with viewing direction $\mathbf{d}$, a neural network predicts the parameters of the radiance and density distributions. (ii) For each spatial-location, we sample a set of radiance-density pairs from the previous distributions. (iii) This generates a set of different radiance-density trajectories along the ray obtained from a distribution of radiance fields. (iiii) The volume rendering equations are used to estimate the RGB values for each  trajectory and compute the log-likelihood for a given pixel color. During inference, the mean and variance of these samples are used as the model prediction and to quantify its associated uncertainty.} 
  \label{fig:pipeline}
\end{figure*}

\subsection{Learning S-NeRF with Variational Inference}
\label{sec:var_inference}
Given that the explicit computation of the posterior in \Eq{snerf_posterior} is intractable, we employ variational inference \cite{blei2017variational} in order to approximate it. In particular, we define a parametric distribution $q_{\boldsymbol{\theta}}(\mathcal{F})$ approximating the true posterior and optimize its parameters $\boldsymbol{\theta}$ by minimizing the Kullback-Leibler (KL) divergence between both as: 
\begin{align}
    \label{eq:var_snerf}
    &\min_\theta \hspace{1mm} \infdiv{q_{\boldsymbol{\theta}}(\mathcal{F})}{p( \mathcal{F} | \mathcal{T})} \nonumber \\
    &\propto -\mathbb{E}_{q_{\boldsymbol{\theta}}(\mathcal{F})} \log(p( \mathcal{T} | \mathcal{F}) ) + \mathbb{E}_{q_{\boldsymbol{\theta}}(\mathcal{F})} \log \bigg( \frac{q_{\boldsymbol{\theta}}(\mathcal{F})}{p(\mathcal{F})} \bigg)
\end{align}

Intuitively, the first term in \Eq{var_snerf} measures the expected training set likelihood over the radiance field distribution $q_{\boldsymbol{\theta}}(\mathcal{F})$. On the other hand, the second term measures the KL divergence between the approximate posterior  and the prior distribution $p(\mathcal{F})$. In the following, we detail how S-NeRF addresses this optimization problem.

\subsubsection{Modelling the approximate posterior}
\label{sec:sample}
In order to make the approximate posterior $q_\theta(\mathcal{F})$ tractable, we define it as a fully-factorized distribution:
\begin{equation}
    \label{eq:apost_definition}
    q_{\boldsymbol{\theta}}(\mathcal{F}) = \prod_{\mathbf{x} \in \mathbb{R}^3} \prod_{\mathbf{d} \in \mathbb{R}^2} q_{\boldsymbol{\theta}}(\mathbf{r}|\mathbf{x},\mathbf{d}) q_{\boldsymbol{\theta}}(\alpha|\mathbf{x})
\end{equation}
where we assume that the density and radiance are independent variables given any location-view pair $(\mathbf{x},\mathbf{d})$. In particular, S-NeRF models $q_{\boldsymbol{\theta}}(\mathcal{F})$ with a neural network defining a function $f_{\boldsymbol{\theta}}(\mathbf{x},\mathbf{d})= \{\mathbf{\mu}_\mathbf{r},\mathbf{\sigma}_
\mathbf{r},{\mu}_\mathbf{\alpha},{\sigma}_
{\alpha} \}$,
where $\mathbf{\mu}_\mathbf{r} \in \mathbb{R}^{3}$ and $\mathbf{\sigma}_
\mathbf{r} \in \mathbb{R}^{+}$ are a mean and standard deviation defining the distribution  $q_{\boldsymbol{\theta}}(\mathbf{r}|\mathbf{x},\mathbf{d})$. Similarly, ${\mu}_\mathbf{\alpha} \in \mathbb{R}$ and ${\sigma}_
{\alpha} \in \mathbb{R}^+$ define the density distribution $q_{\boldsymbol{\theta}}(\alpha|\mathbf{x})$. 

Given that the radiance values need to be bounded between $0$ and $1$, we use a logistic normal distribution \cite{logitnormal} for each RGB channel $r^c$ independently. In this manner, $r^c$ is a random variable defined by:
\begin{equation}
\label{eq:rad_sampling}
r^c = \text{Sigmoid}(\hat{r}^c) \hspace{2mm} \text{;} \hspace{2mm} \hat{r}^c \sim \mathcal{N}(\hat{r}^c | \mu_{r^c},\sigma_{r^c})
\end{equation}
resulting from applying a sigmoid function to a Gaussian variable $\mathcal{N}$ with mean $\mu_{r^c}$ and standard deviation $\sigma_{r^c}$. Similarly, the support for the density distribution  $q_{\boldsymbol{\theta}}(\alpha|\mathbf{x},\mathbf{d})$ needs to be positive. Therefore, we model $\alpha$ as a random variable following a rectified normal distribution \cite{rectifiednormal}:  
\begin{equation}
\label{eq:den_sampling}
\alpha = \psi(\hat{\alpha}) \hspace{2mm} \text{;} \hspace{2mm} \hat{\alpha} \sim \mathcal{N}(\hat{\alpha} | \mu_{\alpha},\sigma_{\alpha})
\end{equation}
where $\psi(\cdot)=\textrm{max}(0,\cdot)$ is as a rectified linear unit that sets all the negative values to $0$.

\begin{figure}[t]
    \centering
    \includegraphics[width=1\linewidth,trim={25 0 0 0},clip]{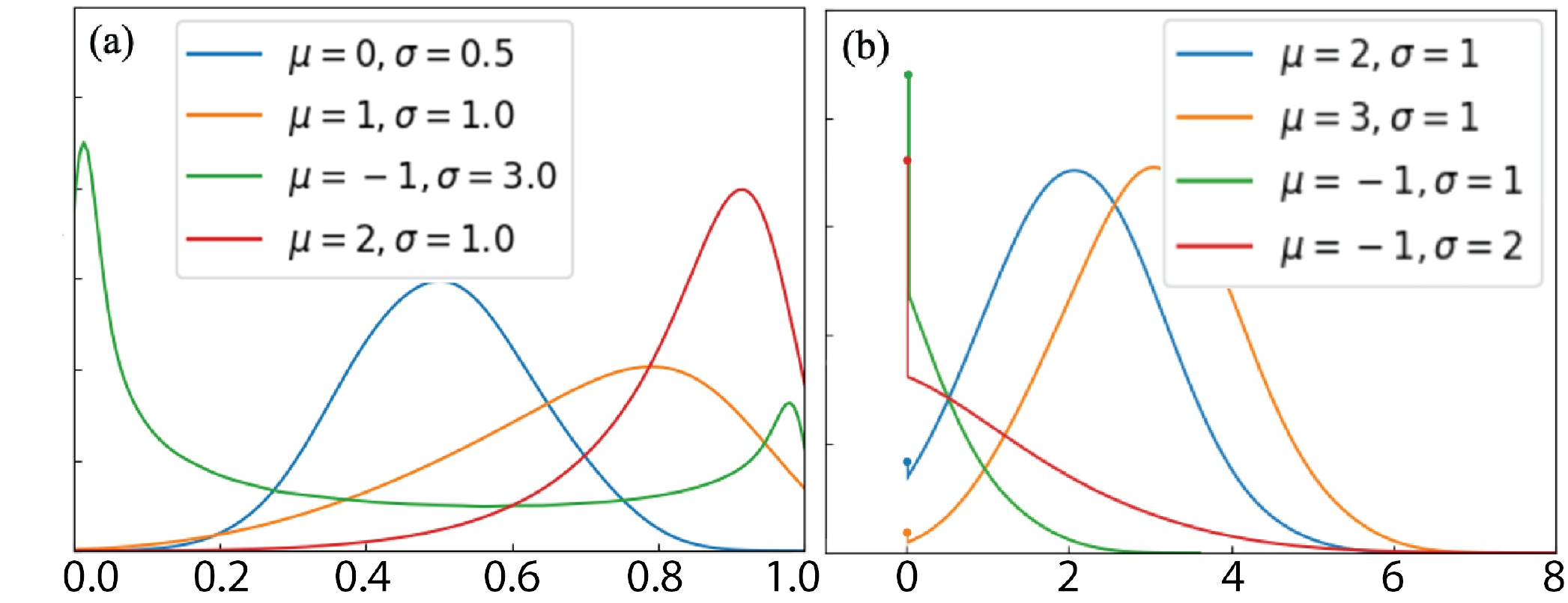}
    \caption{\textbf{Probability Density Functions} for (a) logistic normal (b) and rectified normal distributions with different mean and variance parameters. In S-NeRF the logistic-Normal is used to model the distribution over the radiance values given that its support is bounded between 0 and 1. The volume density distribution is defined by the rectified normal whose support is $[0,\infty]$.}
    \label{fig:dist_illustration}
    \vspace{-3mm}
\end{figure}

Both distributions are illustrated in \fig{dist_illustration} and have a tractable analytical form. Additionally, it is easy to sample from them by applying the Sigmoid or ReLU operation to a variable obtained from a normal distribution with parameters $\{\mathbf{\mu}_\mathbf{r},\mathbf{\sigma}_
\mathbf{r}\}$ and $\{{\mu}_\mathbf{\alpha},{\sigma}_
{\alpha} \}$, respectively. See \app{appendix_distribution_def} for more details.

\subsubsection{Computing the log-likelihood}
\label{sec:log-likelihood}
In the following, we introduce how S-NeRF computes the likelihood term in \Eq{var_snerf}. Firstly, note that given the variational posterior $q_{\boldsymbol{\theta}}(\mathcal{F})$ and the training set $\mathcal{T}$, $\mathbb{E}_{q_{\boldsymbol{\theta}}(\mathcal{F})} \log(p(\mathcal{T}|\mathcal{F}))$ is equivalent to:
\begin{align}
\frac{1}{N} \sum_{n=1}^N \mathbb{E}_{q_{\boldsymbol{\theta}}(\mathbf{r}^n_{t_s:t_f}, \alpha^n_{t_s:t_f} | \mathbf{x}^n_{t_s:t_f},\mathbf{d}^n)} \log(p(\mathbf{c}^n | \mathbf{r}^n_{t_s:t_f},\alpha^n_{t_s:t_f}))
\label{eq:loglikelihood}
\end{align}
where: \textit{(i)} $\mathbf{x}^n_{t_s:t_f}=\{\mathbf{o^n} + t \mathbf{d}^n: t \in [t_s,t_f]\}$ is the set of 3D coordinates along a ray with direction  $\mathbf{d}$ and origin $\mathbf{x}^n_{t_s}$, \textit{(ii)} $\{\mathbf{r}^n_{t_s:t_f},\alpha^n_{t_s:t_f}\}$ is a set of radiance-density pairs for each ray position and \textit{(iii)} $p(\mathbf{c}^n | \mathbf{r}^n_{t_s:t_f},\alpha^n_{t_s:t_f})$ is the probability of the pixel color $\mathbf{c}^n$ given the radiance and density values accumulated along the ray. The latter probability  is 
defined similarly to standard NeRF, where we assume that $\mathbf{c}^n$ follows a normal distribution with a mean defined by applying the volumetric rendering function \Eq{vol_rendering} to the radiance-density trajectory  $\{\mathbf{r}^n_{t_s:t_f},\alpha^n_{t_s:t_f}\}$ along the ray.

Given previous definitions, \Eq{loglikelihood} can be computed using a Monte-Carlo approximation:
\begin{equation}
\label{eq:mc_likelihood}
    \frac{1}{K} \sum_{k=1}^K \log(p(\mathbf{c}^{n} | \mathbf{r}^{nk}_{t_s:t_f},\alpha^{nk}_{t_s:t_f}))
\end{equation}
where each $\mathbf{r}^{nk}_t$ is a sample from the radiance distribution $q_{\boldsymbol{\theta}}(\mathbf{r} | \mathbf{x}^n_t,\mathbf{d}^n)$. These samples can be generated using \Eq{rad_sampling} with parameters $\{\mu_{r},\sigma_r\}$ obtained by evaluating the network $f_{\boldsymbol{\theta}}(\mathbf{x}^n_t,\mathbf{d}^n)$. 
Similarly, $\alpha^{nk}_t$ is a sample from the volume density distribution using \Eq{den_sampling} with mean and variance parameters also defined by the network output. An illustration of the whole process is provided in \fig{pipeline}. The introduced strategy is used during training to compute the log-likelihood and apply stochastic gradient descent to optimize the parameters ${\boldsymbol{\theta}}$. This is possible by using the reparametrization-trick \cite{Reparameterizationtrick} to back-propagate the gradients through the generated samples $\mathbf{r}^{nk}_t$ and $\alpha^{nk}_t$. See \app{reparametrization_trick} for a more detailed explanation.

\subsubsection{Estimating the posterior-prior KL divergence}
\label{sec:priorKL}
As previously discussed, the KL term in \Eq{var_snerf} measures the difference between the approximated posterior over the radiance fields $\mathcal{F}$ learned by S-NeRF and a prior distribution. Similarly to the definition of $q_{\boldsymbol{\theta}}(\mathcal{F})$ in \Eq{apost_definition}, we model the prior $p(\mathcal{F})=\prod_{\mathbf{x} \in \mathbb{R}^3} \prod_{\mathbf{d} \in \mathbb{R}^2} p(\mathbf{r}) p(\alpha)$ as a fully-factorized distribution,

where the radiance and density priors are assumed to be the same for all the spatial-locations in the scene. Concretely, $p(\mathbf{r})$ is again modelled with a logistic normal distribution as in the case of $q_{\boldsymbol{\theta}}(\mathbf{r} | \mathbf{x},\mathbf{d})$. In this case, however, the mean parameter is optimized during training and its variance is fixed to 10. This high value models our knowledge that, without considering any observation, the uncertainty over the radiance values must be high. Analogously, $p(\alpha)$ is modelled with a rectified-normal distribution also with an optimized mean and fixed variance ($\sigma$=10). 

Given previous definitions, the KL term in \Eq{var_snerf} can be expressed as:
\begin{align}
\label{eq:kl_term}
\scriptsize
\mathbb{E}_{q_{\boldsymbol{\theta}}(\mathcal{F})} \log \bigg( \frac{q_{\boldsymbol{\theta}}(\mathcal{F})}{p(\mathcal{F})} \bigg) & = \sum_{\mathbf{x} \in \mathbb{R}^3} \sum_{\mathbf{d} \in \mathbb{R}^2} \infdiv{q_{\boldsymbol{\theta}}(\mathbf{r} | \mathbf{x},\mathbf{d})}{p(\mathbf{r})} \nonumber \\
&+ \sum_{\mathbf{x} \in \mathbb{R}^3} \sum_{\mathbf{d} \in \mathbb{R}^2} \infdiv{q_\alpha(\alpha | \mathbf{x})}{p(\alpha)}
\end{align}
which is equivalent to the sum of the KL divergence between the posterior and prior distributions for all the possible location-view pairs in the scene. During training, \Eq{kl_term} is minimized by sampling random location-view pairs $(\mathbf{x},\mathbf{d})$ in 3D and approximating the radiance and density KL terms using automatic integration. See \app{kl_computation} for more details. 

In our preliminary experiments, we observed that it is also beneficial to consider a different prior distributions for location-view pairs belonging to the rays traced to estimate the pixel-color likelihood in \Eq{loglikelihood}. The reason is that, in these locations, we know that the provided observations are reducing the uncertainty about the radiance-density. Therefore, setting a prior with a high variance for the distributions contradicts this prior knowledge. For these reasons, we also compute the KL term for the spatial-locations sampled along these rays but, in this case, the variances defining the prior distributions $p(\mathbf{r})$ and $p(\alpha)$ are not fix but also optimized during training. A pseudo-algorithm summarizing the learning process of S-NeRF  is provided in \app{pseudocode}. 

\begin{table*}[t]
\centering
\resizebox{0.9\textwidth}{!}{
\begin{tabular}{lcccccccccc}
\multicolumn{1}{c}{} &
   &
  \multicolumn{4}{c}{\textbf{Neg. Log. Likelihood (NLL) $\downarrow$}} &
   &
  \multicolumn{4}{c}{\textbf{MSE-Uncertainty Correlation $\uparrow$}} \\ \cline{3-6} \cline{8-11} 
\multicolumn{1}{c}{} &
   &
  \textbf{MC-DO~\cite{mc-dropout}} &
  \textbf{D. Ens.~\cite{deepensemble}} &
  \textbf{NeRF-W~\cite{nerfwild}} &
  \textbf{S-NeRF} &
   &
  \textbf{MC-DO~\cite{mc-dropout}} &
  \textbf{D. Ens.~\cite{deepensemble}} &
  \textbf{NeRF-W~\cite{nerfwild}} &
  \textbf{S-NeRF} \\ \cline{1-1} \cline{3-6} \cline{8-11} 
\textbf{Flower}   &  & 4.63 & 1.63 & 1.71 & \textbf{1.27}  &  & 0.38 & 0.59 & 0.49 & \textbf{0.63} \\
\textbf{Frotress} &  & 5.19 & 2.29 & 1.04 & \textbf{-0.03} &  & 0.24 & 0.37 & 0.44 & \textbf{0.55} \\
\textbf{Leaves}   &  & 2.72 & 2.66 & 0.79 & \textbf{0.68}  &  & 0.39 & 0.57 & 0.65 & \textbf{0.73} \\
\textbf{Horns}    &  & 4.18 & 2.17 & 0.78 & \textbf{0.60}  &  & 0.43 & 0.50 & 0.50 & \textbf{0.70} \\
\textbf{Trex}     &  & 4.10 & 2.28 & 1.91 & \textbf{1.37}  &  & 0.42 & 0.53 & 0.66 & \textbf{0.68} \\
\textbf{Fern}     &  & 4.90 & 2.47 & 2.16 & \textbf{2.01}  &  & 0.50 & 0.65 & 0.59 & \textbf{0.69} \\
\textbf{Orchids}  &  & 5.74 & 2.23 & 2.24 & \textbf{1.95}  &  & 0.50 & 0.60 & 0.60 & \textbf{0.65} \\
\textbf{Room} &
   &
  5.06 &
  {\color[HTML]{333333} \textbf{2.13}} &
  4.93 &
  2.35 &
   &
  0.46 &
  0.65 &
  0.38 &
  \textbf{0.74} \\ \cline{1-1} \cline{3-6} \cline{8-11} 
\textbf{Avg.}     &  & 4.57 & 2.23 & 1.95 & \textbf{1.27}  &  & 0.40 & 0.56 & 0.54 & \textbf{0.67}
\end{tabular}}
\vspace{1mm}
\caption{\textbf{Results for the uncertainty estimation metrics obtained by all the evaluated methods on the NeRF LLFF dataset~\cite{nerf}.}}
\label{tab:results}
\vspace{-5mm}
\end{table*}

\begin{table}[t]
\centering
\resizebox{0.99\linewidth}{!}{
\begin{tabular}{
>{\columncolor[HTML]{FFFFFF}}c
>{\columncolor[HTML]{FFFFFF}}c 
>{\columncolor[HTML]{FFFFFF}}c 
>{\columncolor[HTML]{FFFFFF}}c 
>{\columncolor[HTML]{FFFFFF}}c }
\hline
 &
  \textbf{MC-DO~\cite{mc-dropout}} &
  \textbf{D. Ens.~\cite{deepensemble}} &
  \textbf{NeRF-W~\cite{nerfwild}} &
  \textbf{S-NeRF} \\
  \hline
Inference (sec.) &
  \multicolumn{1}{r}{\cellcolor[HTML]{FFFFFF}{\color[HTML]{333333} 38.26}} &
  \multicolumn{1}{r}{\cellcolor[HTML]{FFFFFF}{\color[HTML]{000000} 34.77}} &
  \multicolumn{1}{r}{\cellcolor[HTML]{FFFFFF}7.08} &
  \multicolumn{1}{r}{\cellcolor[HTML]{FFFFFF}\textbf{6.06}}\\
  \hline
\end{tabular}}
\vspace{1mm}
\caption{\textbf{Rendering time} (s) required by the evaluated methods for a single view with resolution 500$\times$400.}
\vspace{-8mm}
\label{tab:time}
\end{table}

\subsection{Inference and Uncertainty Estimation}
\label{sec:synth_unc}

By learning a distribution over radiance fields, S-NeRF is able to quantify the uncertainty associated with rendered views for any given camera pose. For this purpose, we firstly sample a set of $K$ color values $\mathbf{c}^k$ for each pixel in the rendered image. As illustrated in \fig{pipeline}, these values are obtained by applying the volume rendering equation to different radiance-density trajectories $\{\mathbf{r}^k_{t_s:t_f},\alpha^k_{t_s:t_f}\}$ along a ray traced from the pixel coordinates. Intuitively, each of the sampled colors $\mathbf{c}^k$ represents an estimate produced by a single radiance field in the learned distribution. Finally, we treat the mean and variance over the $K$ samples $\mathbf{c}^k$ as the predicted pixel color and its associated uncertainty.

Similar to the case of image rendering, S-NeRF is also able to quantify the uncertainty associated with estimated depth-maps. In this case, we ignore the radiance values and use $K$ trajectories $\{\alpha^k_{t_s:t_f}\}$ obtained by sampling density values along the ray. Then, for each sampled trajectory, we compute the expected termination depth of the ray as in ~\cite{nerf}. In this way, we can get $K$ samples for each pixel in the depth maps. The mean and variance of these samples correspond to the estimated depth and its uncertainty.

\section{Experiments}
\subsection{Experimental setup}
\label{sec:exp_setup}

\mypar{Datasets} We conduct our experiments over the LLFF benchmark dataset introduced in~\cite{nerf}. It contains multiple views with calibrated camera poses for 8 different scenes including indoor (\textit{Horns, Trex, Room, Fortress, Fern}) and outdoor environments (\textit{Flower, Leaves, Orchids}). Given that our goal is to evaluate the reliability of the quantified uncertainty, we use a more reduced number of scene views during training compared to the experimental setups used in the original paper. The rationale is the following: in low-data regimes, uncertainty estimation is of particular importance given that the model should be able to identify the parts of the scene that are not covered by the training. In these cases, the model is expected to automatically assign a high uncertainty to these regions. Motivated by this observation, we randomly choose only a $\sim 20\%$ of the total views for training and use the rest for testing. 

\mypar{Baselines} As discussed in \sect{related}, only NeRF-W~\cite{nerfwild} has attempted to quantify uncertainty in Neural Radiance Fields. For this reason, we also compare S-NeRF with state-of-the-art approaches that have been proposed in other domains for the same purpose. In particular, we consider MC-dropout~\cite{mc-dropout} and Deep-Ensembles~\cite{deepensemble}. In the first case, we add a dropout layer after each odd layer in the network to sample multiple outputs using random dropout configurations. Considering a trade-off between computation and performance, we use five samples in our experiments and compute their variance as the uncertainty value.

On the other hand, in Deep-Ensembles we train and evaluate five different NeRF models in parallel. Again, the variance of their outputs is used as the uncertainty associated with the prediction. Finally, we also compare S-NeRF with the proposed strategy used in NeRF-W~\cite{nerfwild} for uncertainty estimation. Given that there are no variable illumination or moving objects in the scene of the evaluated datasets, we remove the latent embedding component of their approach and keep only the uncertainty estimation layers.

\mypar{Evaluation Metrics} Previous works typically evaluate the rendered novel-views using image-quality metrics such as PSNR, SSIM, and LPIPS. However, these validation criteria are not informative in our context given that we aim to measure the reliability of the uncertainty estimates. For this reason, we use two alternative metrics: the negative log-likelihood (NLL) and the correlation between the Mean Squared Error (MSE) and the obtained uncertainty values. The use of the NLL is motivated by the observation that all the evaluated methods provide uncertainty estimations based on a predicted variance for each estimated pixel color. In this manner, we can compute the NLL for each pixel by computing the probability of the ground-truth given a Gaussian distribution with mean equivalent to the estimated color and variance equal to the predicted uncertainty. More intuitively, this metric measures the average MSE error with respect to the color gound-truth weighted by the model confidence associated with each pixel color. 
In the second metric, we compute the correlation between the MSE error for each pixel and the estimated uncertainty values. Note that this correlation will be better if the model assigns higher uncertainty to estimations that are more likely to be inaccurate. Therefore, this metric indicates whether the uncertainty estimates can be used as a predictive value for the expected error in real scenarios where no ground-truth is available.

\mypar{Implementation details}
To implement the different compared baselines, we use the same network architecture, hyper-parameters and optimization process employed in the original NeRF paper \footnote{https://github.com/bmild/nerf}. For S-NeRF implementation, we also use the same architecture to implement the function $f_{\boldsymbol{\theta}}(\mathbf{x},\mathbf{d})$. The only introduced modification is in the last layer, where we double the number of outputs to account for the mean and variance parameters of the density and radiance distributions. During training, we uniformly sample 128 spatial-locations across each ray. Then, for each location, we sample $K=128$ radiance-density pairs $\mathbf{r}^{k}_{t_s:t_f},\alpha^{k}_{t_s:t_f}$ from the distributions defined by the ouput parameters. Finally, to compute the volume rendering formula in \Eq{vol_rendering}, we approximate its integral using the trapezoidal rule (detailed in \app{trapezoidal_rule}). In our preliminary experiments, this integration method showed better stability than the original alpha-compositing used in standard NeRF.

\begin{figure*}[t]
    \centering
    \includegraphics[width=0.96\textwidth]{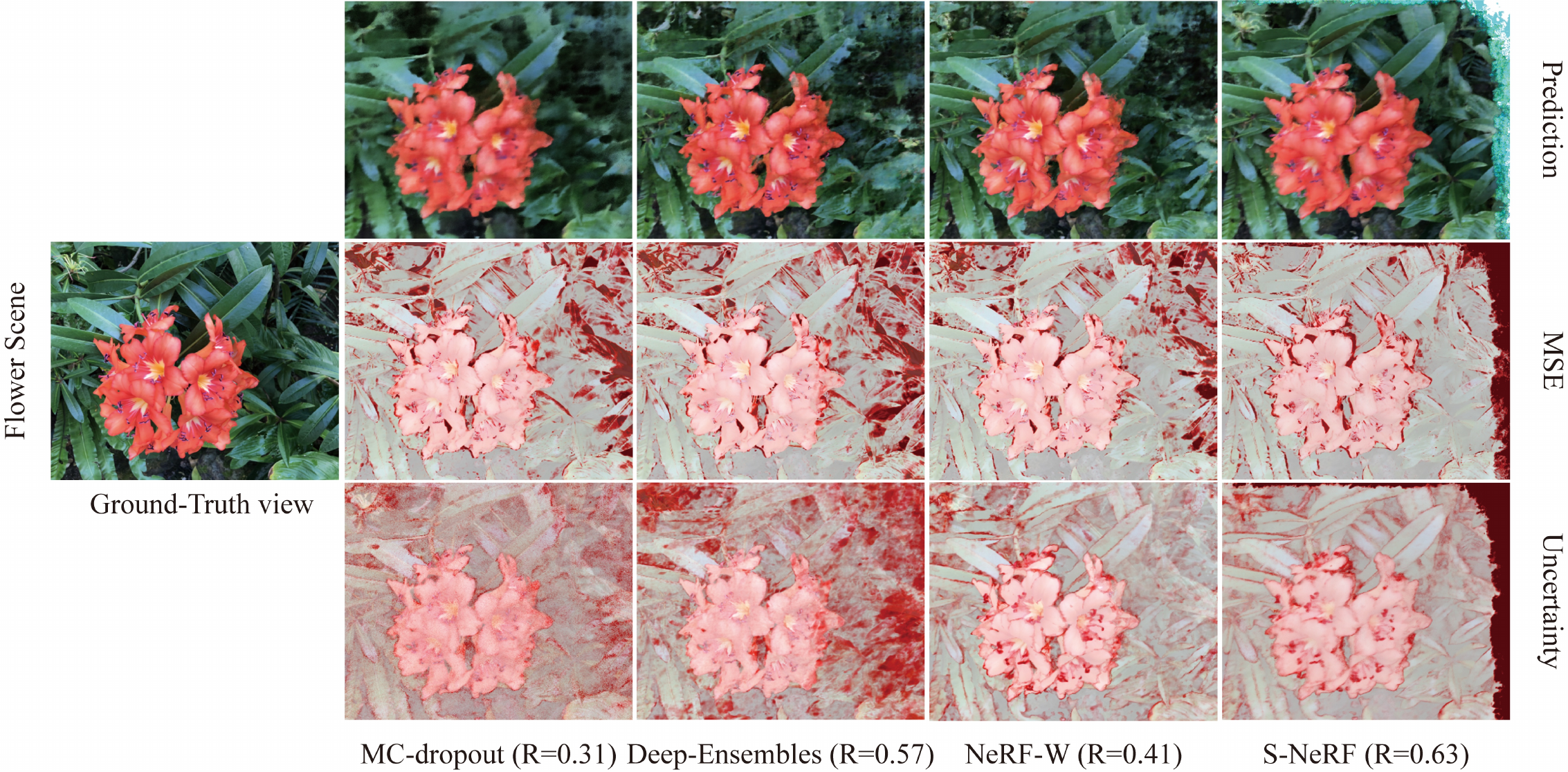}
    \\[\smallskipamount]
    \caption{\textbf{Qualitative results obtained by the evaluated methods}. Rendered view, Mean-Squared-Error with the ground-truth image, and estimated uncertainty are shown respectively from the first to the third row. Compared to other approaches, S-NeRF produces uncertainty estimates that are more correlated with the predictive error (R=0.63). Additionally, note that the right part of the image corresponds to a region not covered by the training views. As a consequence, all the methods obtain a high estimation error in the pixels belonging to this area. S-NeRF, however, is the only method that is able to correctly assign high uncertainty values to these pixels.}
    \label{fig:results}
    \vspace{-3mm}
\end{figure*}

\subsection{Uncertainty estimation in novel-view synthesis}

\noindent\textbf{Quantitative results} of S-NeRF and the other evaluated methods can be found in Table \ref{tab:results}.
As we can observe, our method outperforms all the previous approaches across all the scenes and metrics. In particular, S-NeRF improves over the previous state-of-the-art with an average decrease of 35\% for NLL and with more than 10\% increased MSE-Uncertainty correlation. 
The better results obtained by our approach can be explained by the following reasons. Firstly, the quality of the uncertainty estimates provided by Deep-Ensembles and MC-Dropout typically increases when more models in the ensemble or dropout samples are used, respectively. In our experiments, we have limited this number to $5$ which can partially explain the worse results of MC-Dropout and Deep-Ensembles compared to S-NeRF. While increasing the number of model evaluations in these approaches could improve their performance, this strategy is not practical for NeRF given that the rendering time also grows dramatically. This can be seen in \tab{time}, where we report the rendering time for a single scene required by the different methods. Note that MC-Dropout and Deep-Ensembles  increase the computational complexity of NeRF by a factor of $5$. Whereas NeRF-W has a similar computational complexity to S-NeRF, our method also obtains significantly improved results in all metrics. As discussed in \sect{related}, NeRF-W treats the variance for each pixel as an additional value rendered in the same manner as pixel RGB intensities. This strategy is not theoretically founded and can lead to sub-optimal results. In contrast, our S-NeRF obtains the uncertainty estimates by sampling multiple color values from the posterior distribution over the radiance fields modeling the scene. As we have shown empirically, this strategy produces more accurate uncertainty estimates without increasing the rendering time.

\mypar{Qualitative results} To give more insights on the previous experiments, Figure~\ref{fig:results} shows an example of qualitative results produced by the evaluated models on a testing view. It is important to notice that the right part of the rendered image corresponds to a region that is not covered by the training views used to learn the model. Therefore, we expect to obtain high error and uncertainty estimates in the pixels belonging to this area. As can be observed, the uncertainty values estimated by MC-dropout, Deep-Ensembles and NeRF-W are poorly correlated with their predictive error. As expected, the MSE is high in the right image region that was not observed in the training set views. However, the corresponding uncertainty values provided by these methods are low. In contrast, S-NeRF is able to assign a high uncertainty to the pixels belonging to the scene region that was not covered by the training views. The ability of our model to identify these regions can be explained by the the high-variance defining our prior distributions. Concretely, note that the minimized KL divergence in \Eq{kl_term} forces the learned posterior distribution to resemble this prior when no spatial-locations are observed. As a consequence, the rendered parts of the scene which were not covered by the training views will have an associated high uncertainty. In the following, we conduct an ablation study in order to analyse effect of the prior distributions.

\begin{table}[t]
\centering
\resizebox{.5\textwidth}{!}{
\begin{tabular}{llccclccc}
\multicolumn{1}{c}{} &
   &
  \multicolumn{3}{c}{\textbf{Neg. Log. Likelihood(NLL) $\downarrow$}} &
   &
  \multicolumn{3}{c}{\textbf{MSE-Uncertainty Correlation $\uparrow$}} \\ \cline{3-5} \cline{7-9} 
\multicolumn{1}{c}{} &
   &
  \textbf{w/o KL} &
  \textbf{w/ KL} &
  \textbf{S-NeRF} &
   &
  \textbf{w/o KL} &
  \textbf{w/ KL} &
  \textbf{S-NeRF} \\ \cline{1-1} \cline{3-5} \cline{7-9} 
\cellcolor[HTML]{FFFFFF}\textbf{Flower} &
   &
  \cellcolor[HTML]{FFFFFF}{\color[HTML]{333333} 1.41} &
  \cellcolor[HTML]{FFFFFF}{\color[HTML]{000000} 2.18} &
  \cellcolor[HTML]{FFFFFF}\textbf{1.27} &
   &
  \cellcolor[HTML]{FFFFFF}0.54 &
  \cellcolor[HTML]{FFFFFF}0.51 &
  \cellcolor[HTML]{FFFFFF}\textbf{0.63} \\
\cellcolor[HTML]{FFFFFF}\textbf{Frotress} &
   &
  \cellcolor[HTML]{FFFFFF}1.26 &
  \cellcolor[HTML]{FFFFFF}0.82 &
  \cellcolor[HTML]{FFFFFF}\textbf{-0.03} &
   &
  \cellcolor[HTML]{FFFFFF}0.31 &
  \cellcolor[HTML]{FFFFFF}0.53 &
  \cellcolor[HTML]{FFFFFF}\textbf{0.55} \\
\cellcolor[HTML]{FFFFFF}\textbf{Leaves} &
   &
  \cellcolor[HTML]{FFFFFF}0.98 &
  \cellcolor[HTML]{FFFFFF}0.96 &
  \cellcolor[HTML]{FFFFFF}\textbf{0.68} &
   &
  \cellcolor[HTML]{FFFFFF}0.60 &
  \cellcolor[HTML]{FFFFFF}0.65 &
  \cellcolor[HTML]{FFFFFF}\textbf{0.73} \\
\cellcolor[HTML]{FFFFFF}\textbf{Horns} &
   &
  \cellcolor[HTML]{FFFFFF}3.08 &
  \cellcolor[HTML]{FFFFFF}{\color[HTML]{333333} 0.75} &
  \cellcolor[HTML]{FFFFFF}\textbf{0.60} &
   &
  \cellcolor[HTML]{FFFFFF}0.53 &
  \cellcolor[HTML]{FFFFFF}0.64 &
  \cellcolor[HTML]{FFFFFF}\textbf{0.70} \\
\cellcolor[HTML]{FFFFFF}\textbf{Trex} &
   &
  \cellcolor[HTML]{FFFFFF}1.82 &
  \cellcolor[HTML]{FFFFFF}1.54 &
  \cellcolor[HTML]{FFFFFF}\textbf{1.37} &
   &
  \cellcolor[HTML]{FFFFFF}0.58 &
  \cellcolor[HTML]{FFFFFF}0.66 &
  \cellcolor[HTML]{FFFFFF}\textbf{0.68} \\
\cellcolor[HTML]{FFFFFF}\textbf{Fern} &
   &
  \cellcolor[HTML]{FFFFFF}2.46 &
  \cellcolor[HTML]{FFFFFF}{\color[HTML]{000000} \textbf{1.63}} &
  \cellcolor[HTML]{FFFFFF}2.01 &
   &
  \cellcolor[HTML]{FFFFFF}0.51 &
  \cellcolor[HTML]{FFFFFF}0.66 &
  \cellcolor[HTML]{FFFFFF}\textbf{0.69} \\
\cellcolor[HTML]{FFFFFF}\textbf{Orchids} &
   &
  \cellcolor[HTML]{FFFFFF}4.45 &
  \cellcolor[HTML]{FFFFFF}2.37 &
  \cellcolor[HTML]{FFFFFF}{\color[HTML]{333333} \textbf{1.95}} &
   &
  \cellcolor[HTML]{FFFFFF}0.31 &
  \cellcolor[HTML]{FFFFFF}0.59 &
  \cellcolor[HTML]{FFFFFF}\textbf{0.65} \\
\cellcolor[HTML]{FFFFFF}\textbf{Room} &
   &
  \cellcolor[HTML]{FFFFFF}4.39 &
  \cellcolor[HTML]{FFFFFF}{\color[HTML]{000000} 3.49} &
  \cellcolor[HTML]{FFFFFF}\textbf{2.35} &
   &
  \cellcolor[HTML]{FFFFFF}0.54 &
  \cellcolor[HTML]{FFFFFF}0.65 &
  \cellcolor[HTML]{FFFFFF}\textbf{0.74} \\ \cline{1-1} \cline{3-5} \cline{7-9}
\textbf{Avg.} &
   &
  2.48 &
  1.72 &
  \textbf{1.28} &
   &
  0.49 &
  0.61 &
  \textbf{0.67} \\
\end{tabular}}
\caption{\textbf{The effect of the prior constraints.} Conducting KL divergence constraints by a prior with high fixed variance (w/KL) dramatically improves the model's performance on NLL and MSE-Uncertainty correlation. An additional prior with optimizable variance over the training set further enhances the model's performance as well as stability.}
\label{tab:priorconstraints}
\vspace{-3mm}
\end{table}

\subsection{Analysing the effect of the prior distribution}

The prior distribution defined by S-NeRF allows to identify regions in the scene that are not observed in the training views. Additionally, we also impose a prior with learned variance for the spatial-locations belonging to rays crossing the scene from the observed pixels in the training set (see \sect{priorKL}). To validate the effectiveness of this approach, we have evaluated three different variants of S-NeRF trained by using different strategies: (i) minimizing only the negative log-likelihood term defined in \sect{log-likelihood} and ignoring the KL term, (ii) considering also the KL divergence in \Eq{kl_term} between the prior and posterior distribution, (iii) using the proposed optimization objective where we also minimize the additional prior over the "observed" spatial locations. 

According to the reported results in Table~\ref{tab:priorconstraints}, compared to the case where only the log-likelihood is optimized (w/o KL), minimizing the KL divergence using a high-variance prior significantly improves the performance on both NLL and MSE-Uncertainty correlation (w/KL). As previously discussed, this allows S-NeRF to identify the scene regions which are not observed in the training views and assign a high-uncertainty to the corresponding pixels in rendered images. However, we can also observe a significant improvement for our proposed S-NeRF optimization process, where we impose a learned prior for the observed spatial-locations. The reason is that defining a high-variance prior in these areas can lead to sub-optimal results, given that the KL term hinders the model to minimize the negative log-likelihood. In contrast, this is effectively addressed by applying our proposed prior with learned parameters in observed spatial-locations.

\begin{figure}[t]
    \centering
    \includegraphics[width=\linewidth]{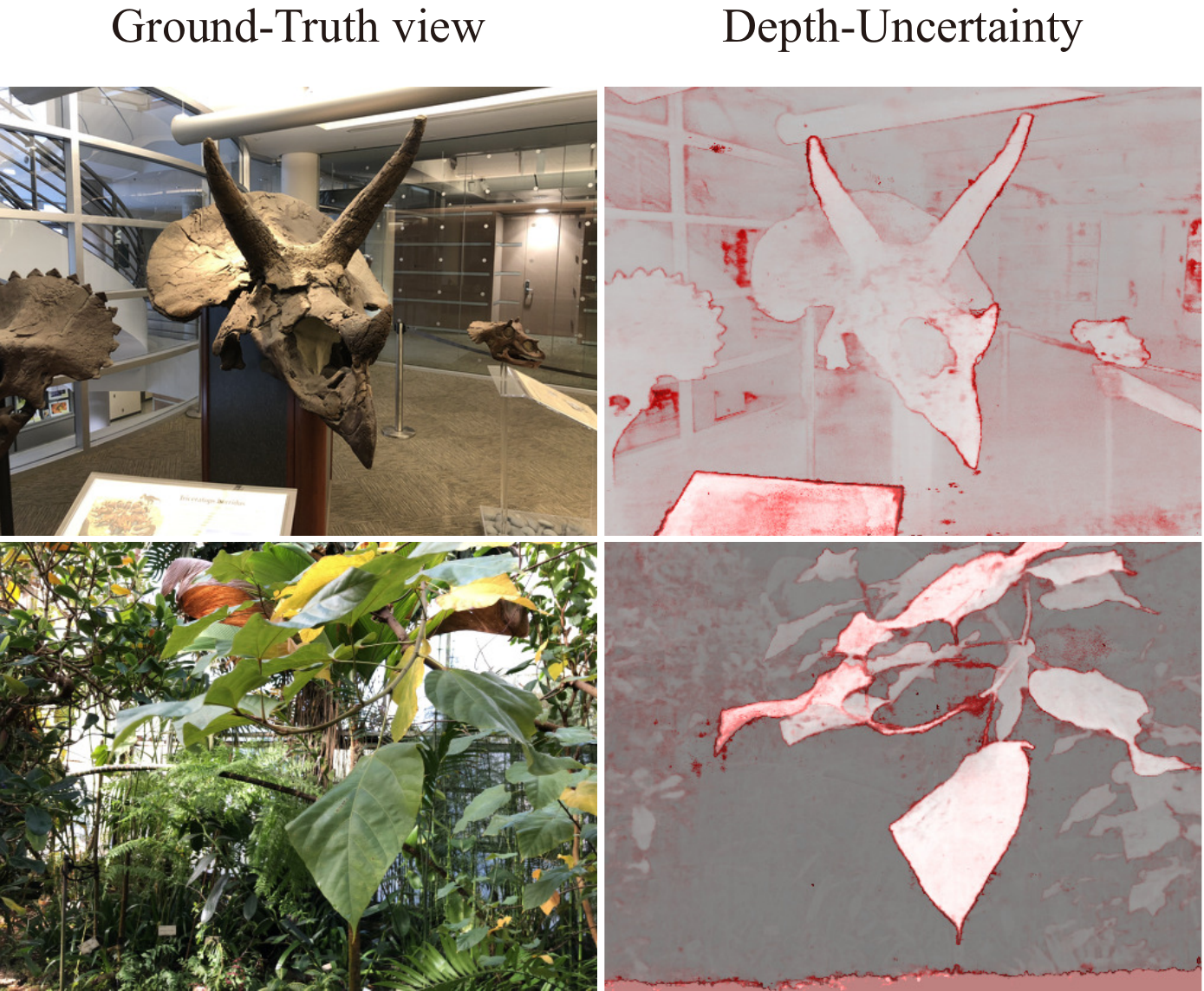}
    \caption{\textbf{Depth-maps and associated uncertainty estimated with S-NeRF.} The high uncertainty at the border of foreground horns and leaves is produced by the discontinuous change of the depth, revealing the low model's confidence on the underlying 3D geometry in those regions.
    }
    \label{fig:depthmap}
    \vspace{-3mm}
\end{figure}

\subsection{Uncertainty estimation in depth-map synthesis}

One of the main advantages of S-NeRF compared to the evaluated methods is that it is also able to quantify the uncertainty associated with the 3D geometry of the scene. In order to illustrate this, \fig{depthmap} shows estimated depth-maps and their associated uncertainty generated for different scenes. By looking at the figure, we can see that our framework can also provide useful information about the model's confidence on the underlying 3D geometry of the scene. For instance, we can observe high uncertainty at the border of foreground objects. This is because this borders correspond to discontinuous changes in the depth-map which produces highly-uncertain estimations. Additionally, we can also observe in the bottom example how S-NeRF is able to assign low confidence values to the depth associated with pixels corresponding to areas of the scene that were not observed in the training set.

\section{Conclusions}

We have presented Stochastic-NeRF, a novel framework to address the problem of uncertainty estimation in  neural volume rendering. The proposed approach is a probabilistic generalization of the original NeRF, which is able to produce uncertainty estimates by modelling a distribution over all the possible radiance fields modelling the scene. Compared to state-of-the-art approaches that can be applied for this problem, we have shown that the proposed method achieves significantly better results without increasing the computational complexity. Additionally, we have also illustrated the ability of S-NeRF to provide uncertainty estimates for different tasks such as depth-map estimation. To conclude, it is also worth mentioning that our formulation is generic and can be combined with any existing or future method based on the NeRF framework in order to incorporate uncertainty estimation in neural 3D representations.

{\small
\bibliographystyle{ieee_fullname}
\bibliography{egbib}
}

\clearpage
\begin{appendices}
\onecolumn
\clearpage
\appendix

\section*{Supplemental Materials} 

\section{Methods}
\label{sec:appendix_methods}

In the following, we provide more technical details about our proposed Stochastic Neural Radiance Fields described in \sect{s_nerf}.

\subsection{Distributions}
\label{sec:appendix_distribution_def}

We present the explicit mathematical expression of the specific distributions used by S-NeRF to model the radiance and density distributions $q(r^c|\mathbf{x},\mathbf{d})$ and $q(\alpha|\mathbf{x})$, respectively. 
Given that the radiance values need to be bounded between 0 and 1, we use a \textbf{logistic normal distribution} for each RGB channel $r^c$ independently. Concretely, its probability density function is defined as:

\begin{equation}
\label{eq:radiance_pdf}
    f(r^c;\mu_{r^c}, \sigma_{r^c}) = \frac{1}{\sigma_{r^c} \sqrt{2\pi}} \frac{1}{r^c(1-r^c)} e^{-\frac{(\text{logit}(r^c)-\mu_{r^c})^2}{2\sigma_{r^c}^2}},
\end{equation}
where $\mu_{r^c}$ and $\sigma_{r^c}$ are the mean and std. deviation of the logit form of variable $r^c$ which is output by the neural network $f_\theta(\mathbf{x},\mathbf{d})$.
Similarly, we model the positive density value $\alpha$ as a random variable following a \textbf{rectified normal distribution}. Its cumulative density function ($\Phi$) and probability density function ($f$) are:

\begin{equation}
\label{eq:density_cdf}
    \Phi(\alpha; \mu_\alpha, \sigma_\alpha) = \frac{1}{\sqrt{2\pi}}\int_{-\infty}^{\alpha} \sigma_\alpha e^{-(\mu_\alpha + \sigma_\alpha t)^2/2} dt
\end{equation}

\begin{equation}
\label{eq:density_pdf}
  f(\alpha;\mu_\alpha, \sigma_\alpha) = 
        \begin{cases}
            \Phi(0;\mu_\alpha,\sigma_\alpha), & \text{if $\alpha \leqslant 0$}\\
            \frac{1}{\sigma_\alpha \sqrt{2\pi}} e^{-\frac{(\alpha-\mu_\alpha)^2}{2\sigma_\alpha^2}} & \text{if $\alpha > 0$}
        \end{cases} 
\end{equation}
where $\mu_\alpha$ and $\sigma_\alpha$ are again the mean and std. deviation of the random variable $\alpha$ output by the S-NeRF network. 

\subsection{Backpropagation through sampling with the reparametrization-trick}
\label{sec:reparametrization_trick}

In the following, we provide a detailed explanation on how to properly sample from the learned distributions and back-propagate their gradients. Sampling directly from the learned distribution for density $\alpha \sim q(\alpha|\mathbf{x})$ and radiance $r^c \sim q(r^c|\mathbf{x},\mathbf{d})$ is not differentiable, which prevents gradient computation of their parameters $\{\mu_{r^c},\sigma_{r^c},\mu_\alpha,\sigma_\alpha \}$ during backpropagation. Inspired by \cite{Reparameterizationtrick}, we introduce a normally distributed variable $\epsilon$ to reparameterize the variables density $\alpha$ and radiance $r^c$. Concretely, we sample density values as $\alpha =\psi(\mu_{\alpha} + \epsilon \sigma_{\alpha})$ in \Eq{den_sampling} and radiance variables as 
$r^c = \text{Sigmoid}(\mu_{r^c} + \epsilon \sigma_{r^c})$ in \Eq{rad_sampling}, where $\epsilon \sim \mathcal{N}(0,1)$ is a unitary gaussian variable with mean $0$ and std. dev.$1$. In this manner, the gradients of the distribution parameters can be computed given that this process is fully-differentiable w.r.t them.

\subsection{The posterior-prior KL divergence}
\label{sec:kl_computation}

As we discuss in \sect{var_inference}, S-NeRF optimization involves the minimization of the KL term in \Eq{var_snerf}. This divergence measures the difference between the approximated posterior learned by S-NeRF and a prior distribution over the radiance fields modelling the scene. Given that computing the KL divergence for all the possible location-view pairs in the scene is intractable, we approximate the sum in \Eq{kl_term} by sampling random 3D spatial-locations in the scene as follows.
Firstly, we define the space bounds at each 3D axis from the captured images of the scene. For instance, we use $x_l$ and $x_r$ to denote the left and right bound of x axis respectively. Then we partition $[x_l,x_r]$ into $N$ evenly-spaced bins and then randomly draw a sample within each bin. After applying this stratified sampling strategy on each axis, we obtain $N\times N\times N$ points paired with random directions. Secondly, for each sampled location-view pair, we use the network $f_{\boldsymbol{\theta}}(\mathbf{x}^n_t,\mathbf{d}^n)$ to compute the posterior distribution parameters $\{\mu_{\mathbf{r}},\sigma_{\mathbf{r}},\mu_\alpha,\sigma_\alpha \}$. Finally, we compute the KL divergence with the prior for the density and radiance variables at each spatial-location. Given that the logistic normal and rectified normal distributions ( \sect{appendix_distribution_def}), the KL divergence between the prior and the posterior in both cases has the following explicit expression:

\textbf{Mathematical derivations for \textbf{KL}$(q_\alpha(\alpha | \mathbf{x}) || p(\alpha))$ in Eq. (\ref{eq:kl_term}):}
\begin{equation}
\begin{aligned}
    &\sum_{\mathbf{x} \in \mathbb{R}^3} \sum_{\mathbf{d} \in \mathbb{R}^2} \infdiv{q_\alpha(\alpha | \mathbf{x})}{p(\alpha)}\\
    &= E_{q_\alpha(\alpha|\mathbf{x})}\log(\frac{q_\alpha(\alpha|\mathbf{x})}{p(\alpha)})\\
    &= \int_{-\infty}^{\infty}
    q_\alpha(\alpha|\mathbf{x})
    \log(\frac{q_\alpha(\alpha|\mathbf{x})}{p(\alpha)})d\alpha \\
    &= \int_{-\infty}^{0} q_\alpha(\alpha|\mathbf{x}) \log(\frac{q_\alpha(\alpha|\mathbf{x})}{p(\alpha)})d\alpha +
    \int_{0}^{\infty} q_\alpha(\alpha|\mathbf{x}) \log(\frac{q_\alpha(\alpha|\mathbf{x})}{p(\alpha)})d\alpha \\
    &= q_\alpha(0|\mathbf{x}) \log(\frac{q_\alpha(0|\mathbf{x})}{p(0)}) + \int_{0}^{\infty} q_\alpha(\alpha|\mathbf{x}) \log(\frac{q_\alpha(\alpha|\mathbf{x})}{p(\alpha)})d\alpha\\
    &= \Phi(0;\mu_q,\sigma_q) [\log(\Phi(0;\mu_q,\sigma_q) -  \log \Phi(0;\mu_p,\sigma_p)] + \int_{0}^{\infty} q_\alpha(\alpha|\mathbf{x}) \log(\frac{q_\alpha(\alpha|\mathbf{x})}{p(\alpha)})d\alpha
\end{aligned}
\end{equation}
Where $\Phi(0;\mu_q,\sigma_q)$ and $q_\alpha(\alpha|\mathbf{x}) = f(\alpha;\mu, \sigma)$ are computed by \Eq{density_cdf} and \Eq{density_pdf} respectively. For the fifth equality, note that the density value is bounded to be positive after a rectified linear unit $\psi(\cdot)=\textrm{max}(0,\cdot)$ that sets all the negative values to $0$. We have shown the intuitive graphics in \fig{dist_illustration}. In practice, we use a Monte-Carlo estimator over density variable during optimization to approximate the integration the previous equations. 

\vspace{2mm}
\textbf{Mathematical derivations for \textbf{KL}$(q_{\boldsymbol{\theta}}(\mathbf{r} | \mathbf{x},\mathbf{d}) || p(\mathbf{r}))$ in Eq. (\ref{eq:kl_term}):}
\begin{equation}
\begin{aligned}
    \sum_{\mathbf{x} \in \mathbb{R}^3} \sum_{\mathbf{d} \in \mathbb{R}^2} \infdiv{q_{\boldsymbol{\theta}}(\mathbf{r} | \mathbf{x},\mathbf{d})}{p(\mathbf{r})} \nonumber
    &= \sum_{c=\{r,g,b\}} E_{q_\theta(r^c|\mathbf{x},\mathbf{d})}\log(\frac{q_\theta(r^c|\mathbf{x},\mathbf{d})}{p(r^c)})\\
    &= \sum_{c=\{r,g,b\}} \int_{-\infty}^{\infty} q_\theta(r^c|\mathbf{x},\mathbf{d}) \log(\frac{q_\theta(r^c|\mathbf{x},\mathbf{d})}{p(r^c)})dr^c
\end{aligned}
\end{equation}
Where $q_\theta(r^c|\mathbf{x},\mathbf{d})$ is equal to $f(r^c;\mu, \sigma)$ for each radiance channel $r^c$, described in \Eq{radiance_pdf}. Again, the integration is approximated by a Monte-Carlo estimator over radiance variable.

\subsection{Trapezoidal rule}
\label{sec:trapezoidal_rule}

To estimate the continuous integral in \Eq{vol_rendering}, firstly we follow the original NeRF~\cite{nerf} to use a stratified sampling strategy to sample $N$ spatial-locations along each ray. Concretely, we partition $[t_s,t_f]$ into $N$ evenly-spaced bins and then randomly draw a sample within each bin:

\begin{equation}
    t_i \sim U\left[t_s + \frac{i-1}{N}(t_f-t_s), t_s + \frac{i}{N}(t_f-t_s)\right]
\end{equation}

For these samples, we can utilize our framework to produce $K$ density-radiance pairs $\{\mathbf{r}^{nk}_{t_s:t_f},\alpha^{nk}_{t_s:t_f}\}$ for each spatial-location along each ray. As mentioned in \sect{exp_setup}, we use a different strategy to estimate the continuous integral in \Eq{vol_rendering} compared to S-NeRF. The motivation is that,
in our preliminary experiments, we observed that using an alternative trapezoidal rule \footnote{Rahman, Qazi I.; Schmeisser, Gerhard (December 1990), "Characterization of the speed of convergence of the trapezoidal rule", Numerische Mathematik, 57 (1): 123–138.} to approximate the volume rendering integral is much stable than the traditional alpha compositing used in the original paper~\cite{nerf}. The reason is that the latter can produce extremely large density values when large variances are associated with the sampled distributions. As a consequence, this produces numerical instabilities during optimization. The alternative trapezoidal method used to approximated the aforementioned integral is able to address this limitation and can be expressed as:

\begin{equation} 
\label{eq:tp}
\hat{C}^k(\mathbf{x}_o,\mathbf{d}) = \sum_{i=1}^{N} \frac{1}{2} \left(T_{i-1}^k \mathbf{r}_{i-1}^k \alpha_{i-1}^k + T_{i}^k \mathbf{r}_{i}^k \alpha_{i}^k \right) \delta_i, \hspace{2mm} \text{where} \hspace{2mm} T_i^k = \exp \left(- \sum_{j=1}^{i-1} \frac{1}{2} (\alpha_{j-1}^k + \alpha_j^k) \delta_j \right)
\end{equation}
where $\delta_i = t_i-t_{i-1}$ is the distance between adjacent sampled spatial-locations along the ray. $(\mathbf{r}_{i}^k, \alpha_{i}^k)$ is the $k$th sample of the $K$ density-radiance pairs at the $i$th ray location.

\section{Pseudo-algorithm}
\label{sec:pseudocode}

In Algorithm \ref{alg:pseudocode}, we provide the pseudocode for the learning process of S-NeRF. 

\begin{algorithm}[]
  \caption{Pseudo-algorithm for learning procedure}
  \label{alg:pseudocode}
  \begin{algorithmic}[1]
    \STATE 
    Given N training triplets $(\mathbf{c}^N,\mathbf{x}^N_o,\mathbf{d}^N)$, trace a ray from each camera origin $\mathbf{x}^N_o$ along direction $\mathbf{d}^N$ and sample n points as input pairs $(\mathbf{x}^n_t,\mathbf{d}^n)$
    \vspace{1mm}
    \STATE For each input pair, use $f_{\boldsymbol{\theta}}(\mathbf{x}^n_t,\mathbf{d}^n)$ to output approximated posterior $q_{\boldsymbol{\theta}}(\mathbf{r}|\mathbf{x},\mathbf{d})$ in \Eq{rad_sampling} and $q_{\boldsymbol{\theta}}(\alpha|\mathbf{x})$ in \Eq{den_sampling}
    \vspace{1mm}
    \STATE Sample $K$ radiance-density pairs $\{\mathbf{r}^n_{t_s:t_f},\alpha^n_{t_s:t_f}\}$ from  $q_{\boldsymbol{\theta}}(\mathbf{r}|\mathbf{x},\mathbf{d})$ and $q_{\boldsymbol{\theta}}(\alpha|\mathbf{x})$ for each input pair, and then get $K$ radiance-density trajectories for each ray
    \vspace{1mm}
    \STATE Integrate each trajectory by volume rendering in \Eq{vol_rendering} to get $K$ samples of the pixel color $\mathbf{c}^N_k$
    \STATE Compute the log-likelihood with $\mathbf{c}^N$ and $\mathbf{c}^N_k$ by \Eq{loglikelihood} and \Eq{mc_likelihood}
    \vspace{1mm}
    \STATE Sample from all the scene coordinates uniformly and compute the KL term by \Eq{kl_term}
    \vspace{1mm}
    \STATE Back-propagate and optimize ${\boldsymbol{\theta}}$ and distribution parameters $\{\mu_{r^c},\sigma_{r^c},\mu_\alpha,\sigma_\alpha \}$ by SGD
  \end{algorithmic}
\end{algorithm}

\section{Additional Qualitative Results}

\begin{figure*}[ht]
    \centering
    \includegraphics[width=\textwidth]{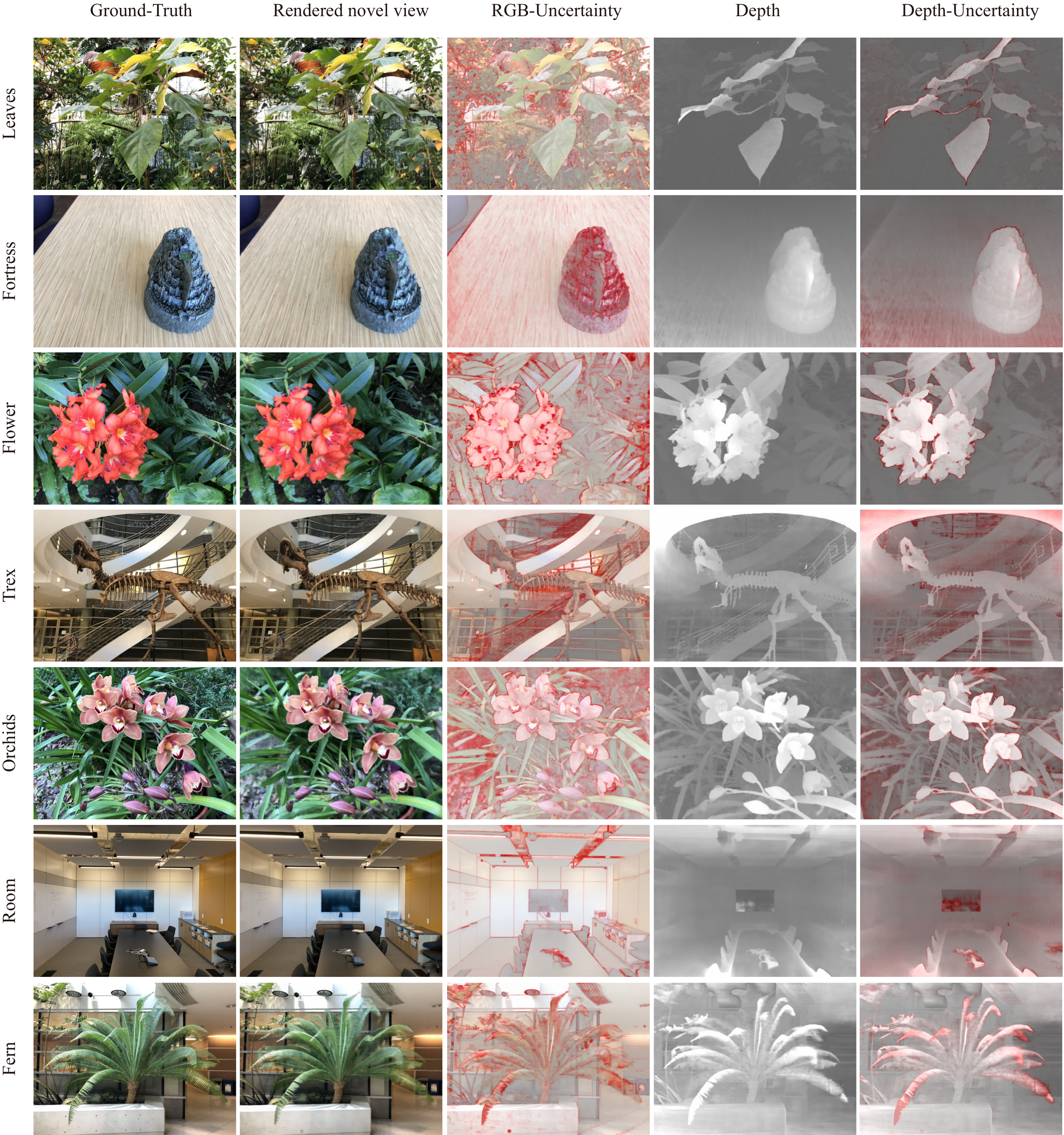}
    \\[\smallskipamount]
    \caption{\textbf{Additional qualitative results obtained by our S-NeRF.}}
    \label{fig:supp_results}
\end{figure*}

In \fig{supp_results}, we show additional qualitative obtained by our S-NeRF results across different scenes in the evaluated dataset. For each scene, we show not only the quantified uncertainty (third column) associated with the rendered novel view (second column), but also the estimated uncertainty (fifth column) associated with the generated depth-map (fourth column).

\end{appendices}

\end{document}